\def\year{2021}\relax
\documentclass[letterpaper]{article} 
\usepackage{aaai21}  
\usepackage{times}  
\usepackage{helvet} 
\usepackage{courier}  
\usepackage[hyphens]{url}  
\usepackage{graphicx} 
\usepackage{natbib}  
\usepackage{caption} 
\usepackage{amsmath}
\usepackage{subcaption}
\usepackage{booktabs}
\usepackage[backref]{hyperref}

\title{Label Confusion Learning to Enhance Text Classification Models}
\author {
        Biyang Guo \thanks{All authors contribute equally and are listed alphabetically.},
        Songqiao Han \footnotemark[1],
        Xiao Han \footnotemark[1],
        Hailiang Huang\footnotemark[1]\thanks{Corresponding author.},
        Ting Lu\footnotemark[1]\\
}
\affiliations {
    AI Lab, School of Information Management and Engineering, \\Shanghai University of Finance and Economics,
    Shanghai, China, 200433 \\
    guobiyang2020@gmail.com, \{han.songqiao, xiaohan, hlhuang\}@shufe.edu.cn, luting@189.cn
}

\begin{document}
\maketitle
\begin{abstract}
Representing a true label as a one-hot vector is a common practice in training text classification models. However, the one-hot representation may not adequately reflect the relation between the instances and labels, as labels are often not completely independent and instances may relate to multiple labels in practice. The inadequate one-hot representations tend to train the model to be over-confident, which may result in arbitrary prediction and model overfitting, especially for confused datasets (datasets with very similar labels) or noisy datasets (datasets with labeling errors). While training models with label smoothing (LS) can ease this problem in some degree, it still fails to capture the realistic relation among labels. In this paper, we propose a novel Label Confusion Model (LCM) as an enhancement component to current popular text classification models. LCM can learn label confusion to capture semantic overlap among labels by calculating the similarity between instances and labels during training and generate a better label distribution to replace the original one-hot label vector, thus improving the final classification performance. Extensive experiments on five text classification benchmark datasets reveal the effectiveness of LCM for several widely used deep learning classification models. Further experiments also verify that LCM is especially helpful for confused or noisy datasets and superior to the label smoothing method. 

\end{abstract}

\section{Introduction}
Text classification is one of the fundamental tasks in natural language processing (NLP) with wide applications such as sentiment analysis, news filtering, spam detection and intent recognition. Plenty of algorithms, especially deep learning-based methods, have been applied successfully in text classification, including recurrent neural networks (RNN),  convolutional networks (CNN)  \cite{kim2014convolutional}. More recently, large pre-training language models such as ELMO \cite{peters2018deep}, BERT \cite{devlin2018bert}, Xlnet \cite{yang2019xlnet} and so on have also shown their outstanding performance in all kinds of NLP tasks, including text classification. 

Although numerous deep learning models have shown their success in text classification problems, they all share the same learning paradigm: a deep model for text representation, a simple classifier to predict the label distribution and a cross-entropy loss between the predicted probability distribution and the one-hot label vector. However, this learning paradigm have at least two problems: (1) In general text classification tasks, one-hot label representation is based on the assumption that all categories are independent with each other. But in real scenarios, labels are often not completely independent and instances may relate to multiple labels, especially for the confused datasets that have similar labels. As a result, simply representing the true label by a one-hot vector fails to take the relations between instances and labels into account, which further limits the learning ability of current deep learning models. (2) The success of deep learning models heavily relies on large annotated data, noisy data with labeling errors will severely diminish the classification performance, but it is inevitable in human-annotated datasets. Training with one-hot label representation is particularly vulnerable to mislabeled samples as full probability is assigned to a wrong category. In brief, the limitation of current learning paradigm will lead to  confusion in prediction that the model is hard to distinguish some labels, which we refer as label confusion problem (LCP). A label smoothing (LS) method is proposed to remedy the inefficiency of one-hot vector labeling \cite{muller2019does}, however, it still fails to capture the realistic relation among labels, therefore not enough the solve the problem.
    
In this work, we propose a novel Label Confusion Model (LCM) as an enhancement component to current deep learning text classification models and make the model stronger to cope with label confusion problem. In particular, LCM learns the representations of labels and calculates their semantic similarity with input text representations to estimate their dependency, which is then transferred to a label confusion distribution (LCD). After that, the original one-hot label vector is added to the LCD  with a controlling parameter and normalized by a softmax function to generate a simulated label distribution (SLD). We use the obtained SLD to replace the one-hot label vector and supervise the training of model training. With the help of LCM, a deep model not only capture
s the relations between instances and labels, but also learns the overlaps among different labels, thus, performs better in text classification tasks. We conclude our contributions as follows:
\begin{itemize}
\item We propose a novel label confusion model (LCM) as an effective enhancement component for text classification models, which models the relations between instances and labels to cope with LCP problems. In addition, LCM is only used during training and doesn't change the original model structure, which means LCM can improve the performance without extra computation cost in prediction procedure.
\item Extensive experiments on five benchmark datasets (both in English and Chinese) illustrate the effectiveness of LCM on three widely used deep learning structures: LSTM, CNN and BERT.  Experiments also verified its advantage over label smoothing method. 
\item We construct four datasets with different confusion degree, and four datasets with different proportion of noise. Experiments on these datasets demonstrate that LCM is especially helpful for confused or noisy datasets and superior to the label smoothing method (LS) to a large degree.
\end{itemize}

\section{Related Work}
\subsection{Text Classification With Deep Learning}
Deep learning models have been widely use in natural language processing, including text classification problems. The studies of deep text representations can be categorized into two groups. One is focusing on the word embeddings\cite{mikolov2013distributed, le2014distributed, pennington2014glove}. Another group mainly study the deep learning structures that can learn better text representations. Typical deep structures include recurrent neural networks (RNNs) based long short-term memory (LSTM) \cite{LSTM,liu2016recurrent,wang2018topic}, convolutional neural networks (CNN) \cite{kalchbrenner2014convolutional,kim2014convolutional, zhang2015character,shen2017deconvolutional} and context-dependent language models like BERT \cite{devlin2018bert}.The reason why deep learning methods have become so popular is their ability to learn sophisticated semantic representations from text, which are much richer than hand-crafted features. 

\subsection{Label Smoothing}
Label smoothing (LS) is first proposed in image classification tasks as a regularization technique to prevent the model from predicting the training examples too confidently, and has been used in many state-of-the-art models, including image classification \cite{szegedy2016rethinking,zoph2018learning}, language translation \cite{vaswani2017attention} and speech recognition \cite{chorowski2016towards}. LS improves model accuracy by computing loss not with the ``hard" one-hot targets, but with a weighted mixture of these targets with a uniform noise distribution.

Nevertheless, the label distribution generated form LS cannot reflect the true label distribution for each training sample, since it is obtained by simply adding some noise. The true label distribution should reveal the semantic relation between the instance and each label, and similar labels should have similar degree in the distribution. In nature, label smoothing encourages the model to learn less, rather than learn more accurately of the knowledge in training samples, which may have the risk of underfitting.

\subsection{Label Embedding}
Label embedding is to learn the embeddings of the labels in classification tasks and has been proven to be effective. \cite{zhang2017multi} convert labels into semantic vectors and thereby convert the classification problem into vector matching tasks. Then attention mechanisms are used to jointly learn the embedding of words and labels \cite{wang2018joint}. \cite{yang2018sgm} use label embedding in a sequence generation model for multi-label classification which captures the co-relation between labels. In our work, we also use jointly learn the label embeddings, which can be used to further capture the semantic relation between text and labels.

\subsection{Label Distribution Learning}
Label Distribution Learning (LDL) \cite{LDL} is a novel machine learning paradigm for applications where the overall distribution of labels matters. A label distribution covers a certain number of labels, representing the degree to which each label describes the instance. LDL is proposed for problems where the distribution of labels matters. \cite{LDL,LDL4C} gives out several algorithms for this kind of tasks. However, the true label distribution is hard to obtain for many existing classification tasks such as 20NG (a typical text classification dataset) and MNIST (a typical image classification dataset) \cite{mnist} where we only have a unique label for each sample. In this kind of classification tasks, LDL is not applicable.

\begin{figure*}[t]
\centering
\includegraphics[width=0.8\textwidth]{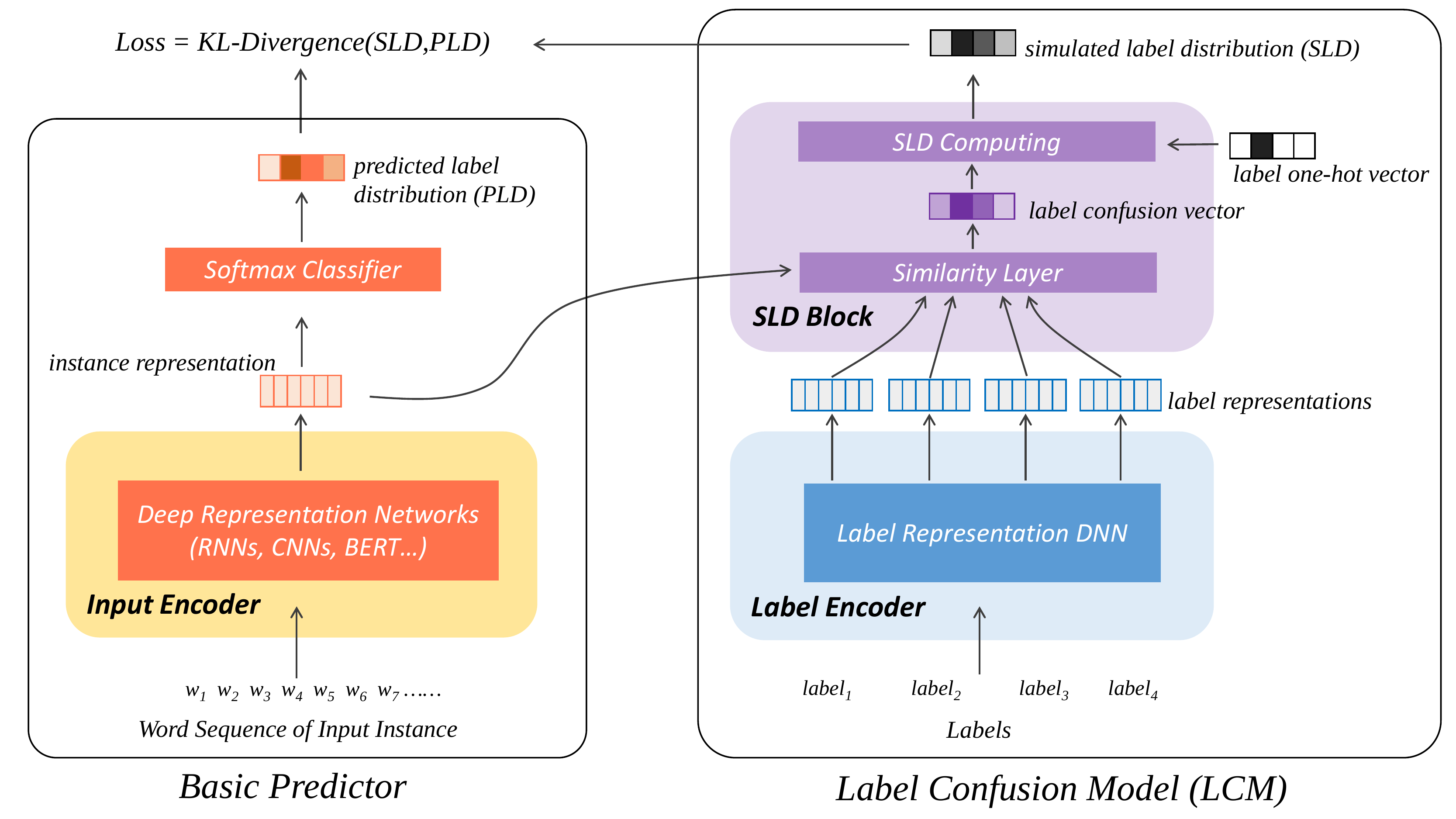} 
\caption{The structure of LCM-based classification model, which is composed of a basic predictor and a label confusion model.}
\label{LCM}
\end{figure*}

\begin{table*}[t]
\small
\centering
\begin{tabular}{l | c c c c c} 
 \toprule[1.5pt]
 \textbf{Models} & \textbf{20NG} & \textbf{AG's News} &  \textbf{DBPedia} &  \textbf{FDCNews} &  \textbf{THUCNews} \\
 \midrule
 LSTM-rand & $0.6822 \pm 0.0334$ & $0.8771 \pm 0.0065$  & $0.9208 \pm 0.0081$ & $0.7726 \pm 0.0181$ & $0.8604 \pm 0.0031$ \\ 

 LSTM-rand\ +\ LCM & $\mathbf{0.7242 \pm 0.0191}$ & $\mathbf{0.8814 \pm 0.0049}$ & $0.9248 \pm 0.0043$ & $\mathbf{0.7988 \pm 0.0059}$ & $\mathbf{0.8655 \pm 0.0027}$ \\ 
 \midrule
 \textit{Improvement} & $4.20\%$ & $0.43\%$ & $0.40\%$ & $2.62\%$ & $0.51\%$ \\
 \midrule[1.0pt]

 LSTM-pre & $0.7416 \pm 0.0070$ & $0.8925 \pm 0.0041$ & $0.9605 \pm 0.0058$ & $0.8298 \pm 0.006$3 & $0.9080 \pm 0.0058$ \\ 
 LSTM-pre\ +\ LCM & $\mathbf{0.7509 \pm 0.0014}$ & $0.8922 \pm 0.0033$ & $\mathbf{0.9643 \pm 0.0046}$ & $\mathbf{0.8352 \pm 0.0050}$ & $\mathbf{0.9126 \pm 0.0051}$ \\ 
 \midrule
 \textit{Improvement} & $0.93\%$ & $-0.03\%$ & $0.35\%$ & $0.54\%$ & $0.46\%$  \\
 \midrule[1.0pt]
 
 CNN-rand & $0.8172 \pm 0.0057$ & $0.8775 \pm 0.0049$ & $0.9560 \pm 0.0020$ & $0.8529 \pm 0.0047$ & $0.8928 \pm 0.0019$ \\ 
 CNN-rand\ +\ LCM & $\mathbf{0.8276 \pm 0.0034}$ & $\mathbf{0.8828 \pm 0.0035}$ &	$\mathbf{0.9606 \pm 0.0016}$ & $\mathbf{0.8583 \pm 0.0022}$ & $\mathbf{0.8990 \pm 0.0022}$ \\ 
 \midrule
 \textit{Improvement} & $1.04\%$ & $0.53\%$ & $0.46\%$ & $0.54\%$ & $0.62\%$  \\
 \midrule[1.0pt]

 CNN-pre & $0.7756 \pm 0.0030$ & $0.8959 \pm 0.0018$ &	$0.9796 \pm 0.0010$ & $0.9023 \pm 0.0018$ & $0.9331 \pm 0.0007$ \\ 
 CNN-pre\ +\ LCM & $0.7713 \pm 0.0047$ & $\mathbf{0.8974 \pm 0.0011}$ &	$\mathbf{0.9811 \pm 0.0013}$ & $0.9008 \pm 0.0024$ & $\mathbf{0.9341 \pm 0.0010}$ \\ 
 \midrule
 \textit{Improvement} & $-0.43\%$ & $0.15\%$ & $0.15\%$ & $-0.15\%$ & $0.10\%$  \\
 \midrule[1.0pt]

 BERT & $0.8853 \pm 0.0043$ & $0.9024 \pm 0.0011$ & $0.9822 \pm 0.0005$ & $0.9488 \pm 0.0042$ & $0.9595 \pm 0.0010$ \\ 
 BERT\ +\ LCM & $\mathbf{0.8896 \pm 0.0034}$ & $\mathbf{0.9042 \pm 0.0008}$ & $\mathbf{0.9834 \pm 0.0006}$ & $\mathbf{0.9544 \pm 0.0034}$ & $\mathbf{0.9607 \pm 0.0011}$ \\ 
 \midrule
 \textit{Improvement} & $0.43\%$ & $0.18\%$ & $0.12\%$ & $0.56\%$ & $0.12\%$  \\
 \bottomrule[1.5pt]
\end{tabular}
\caption{Test Accuracy on different text classification tasks. We report their mean $\pm$ standard deviation. The bold means significant improvement on baseline methods on t-test (p \textless\ 0.1).}
\label{Results}
\end{table*}

\section{Our Approach}
Intuitively, there exists a label distribution which reflect the degree of how each label describes the current instance for most classification tasks. However, in practice, we can only obtain a unique label (in single-label classification) or several labels (in multi-label classification) for samples, rather than the relation degree between the samples and the labels. There isn't a natural and verified way to transfer the one-hot label to a proper distribution, if no statistical information is provided. Though the  theoretical true label distribution cannot be easily achieved, we can still try to simulate it by digging the semantic information behind instances and labels.

Considering that label confusion problem is usually caused by the semantic similarity, we suppose that a label distribution that can reflect the similarity relation between labels can help to train a stronger model and address the label confusion problem. A simple idea is to find the descriptions of each label and calculate the similarity between every two labels. Then for each one-hot label representation, we can use the normalized similarity values to create a label distribution. However, the label distributions got in this way are all the same for instances with the same label, regardless of their content. In fact, even two instances have the same label, their content may be quite different, so their label distribution should also be different. 

Therefore, we should construct the label distribution using the relations between instances and labels, thus the label distribution will dynamically be changing for different instances with the same label. For text classification problems, we can simulate the label distribution by the similarity between the representation of document text and each label. In this way, not only the relations between instances and labels are captured, the dependency among labels is also reflected by these relations. With this basic idea, we designed a label confusion model (LCM) to learn the simulated label distribution (SLD) by calculating the semantic relations between instances and labels. Then the SLD is seen to be the true label distribution and is compared with the predicted distribution to compute the loss by KL-divergence. In the latter part, we will introduce the LCM-based classification model in detail.

\subsection{LCM-based Classification Model}

A LCM-based classification model is composed by two parts: a basic predictor and a label confusion model. The overall structure can be seen in Figure \ref{LCM}. 

The basic predictor is usually constructed by a input encoder (such as RNNs, CNN, BERT) followed by a simple classifier (usually a softmax classifier). The basic predictor can be chosen from all kinds of main stream deep learning based text classifiers. As shown in Figure 1, the text to be classified is inputted to the input decoder to generate the input text representation. Then it will be fed to the softmax classifier to predict the label distribution (PLD):
\begin{align*} 
    v^{(i)} &= f^I(x) = f^I([x_1,x_2,...,x_n]) \\
    &= [v^{(i)}_1,v^{(i)}_2,...,v^{(i)}_n] \\
    y^{(p)} &= softmax(v^{(i)})
\end{align*}
where $f^I$ is the input encoder function, transforming the input sequence $x=[x_1,x_2,...,x_n]$ with length $n$ to the input text representation $v^{(i)}=[v^{(i)}_1,v^{(i)}_2,...,v^{(i)}_n]$ with length $n$ and dimension $d$. $y^{(p)}$ is the predicted label distribution PLD.

The LCM is constructed by a label encoder and a simulated label distribution computing block (SLD Block). The label encoder is a deep neural network (DNN) to generate the label representation matrix. The SLD block is composed of a similarity layer and a SLD computing layer. The similarity layer takes the label representations and the current instance representation as inputs, and computes their similarity values by dot product, then a neural net with softmax activation is applied to get the label confusion distribution (LCD). The LCD captures the dependency among labels by calculating the similarity between instances and labels.  Thereby, the LCD is a dynamic, instance-dependent distribution, which is superior to the distribution that solely considers the similarity among labels, or simply a uniform noise distribution like the way in LS.

Finally, the original one-hot vector is added to the LCD with a controlling parameter $\alpha$, and then normalized by a softmax function the generate the simulated label distribution SLD. The controlling parameter $\alpha$ decides how much of the one-hot vector will be changed by the LCD. The above process can be formulated by:
\begin{align*}
    V^{(l)} 
    &= f^L(l) = f^L([l_1,l_2,...,l_C])\\
    &= [v^{(l)}_1,v^{(l)}_2,...,v^{(l)}_C]\\
    y^{(c)} &= softmax({v^{(i)}}^\top V^{(l)} W + b) \\ 
    y^{(s)} &= softmax(\alpha y^{(t)} + y^{(c)})
\end{align*}
where $f^L$ is the label encoder function to transfer labels $l=[l_1,l_2,...,l_C]$ to label representation matrix $V^{(l)}$, $C$ is the number of categories. $f^L$ in our case is implemented by a embedding lookup layer followed by a DNN, which can be multi-layer perceptron  (MLP), LSTM or attention networks and so on. Note that the order of the label sequence inputted to LCM should be the same with the one-hot targets. $y^{(c)}$ is the LCD and $y^{(s)}$ is the SLD.

The SLD is then be viewed as the new training targets to replace the one-hot vector and supervise the model training. Since the SLD $y^{(s)}$ and the predicted label vector $y^{(p)}$ are both probability distributions, we use the Kullback–Leibler divergence (KL-divergence) \cite{kl} as the loss function to measure their difference:
\begin{align*}
    loss
    &= \textit{KL-divergence}(y^{(s)},y^{(p)}) \\ 
    &= \sum _{c}^{C} y^{(s)}_c \log(\frac{y^{(s)}_c}{y^{(p)}_c})
\end{align*}

By training with LCM, the actual targets the model trying to fit are dynamically changing according to the semantic representation of documents and labels learned by the deep model. The learned simulated label distribution help the model to better represent the instance with different labels, especially for those easily confused samples. SLD is also more robust when facing noisy data cause the probability of the wrong label is allocated to similar labels (often include the right label), thus model can still learn some useful information from mislabeled data.

\begin{figure*}[t]
\centering
     \begin{subfigure}[t]{0.35\textwidth}
         \centering
         \includegraphics[width=\textwidth]{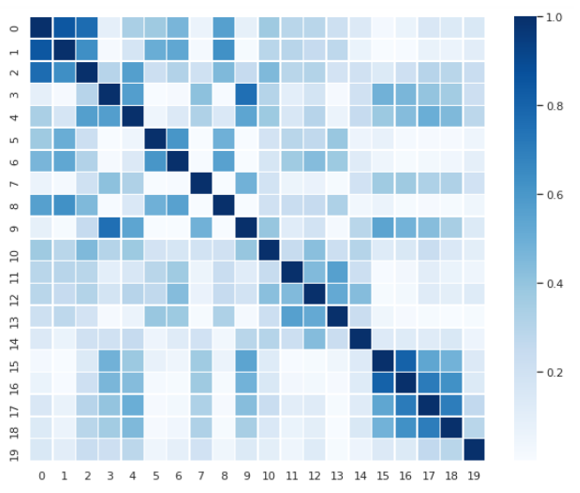}
         \caption{Cosine similarity matrix}
         \label{LabelSimilarityMatrix}
     \end{subfigure}
     \hfill
     \begin{subfigure}[t]{0.55\textwidth}
         \centering
         \includegraphics[width=\textwidth]{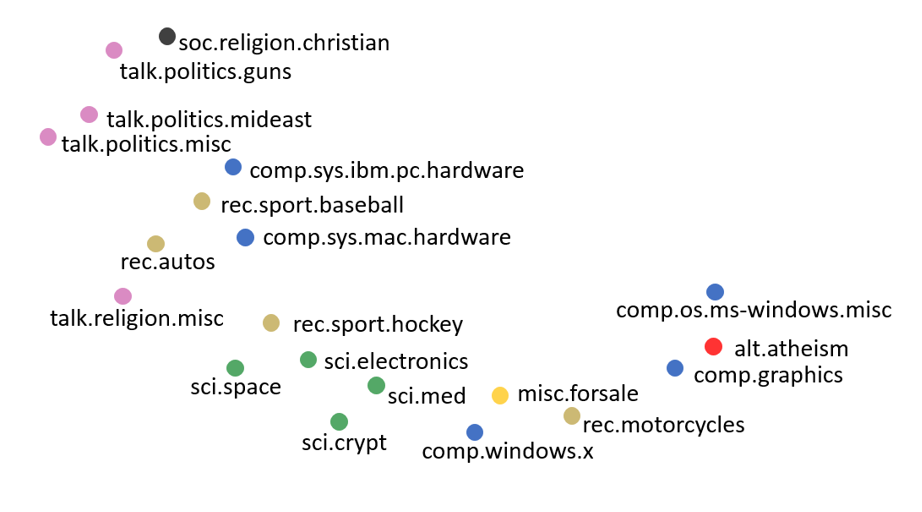}
         \caption{t-SNE visualization of label representations}
         \label{LabelTSNE}
     \end{subfigure}
\caption{Cosine similarity matrix (a) and corresponding t-SNE visualization (b) of label representations of 20NG datasets. The representations are drawn from the embedding layer of LCM. The labels with the same color are of the same label group.}
\label{LabelEmbeddingVis}
\end{figure*}

\begin{figure*}[t]
\centering
     \begin{subfigure}[t]{0.45\textwidth}
         \centering
         \includegraphics[width=\textwidth]{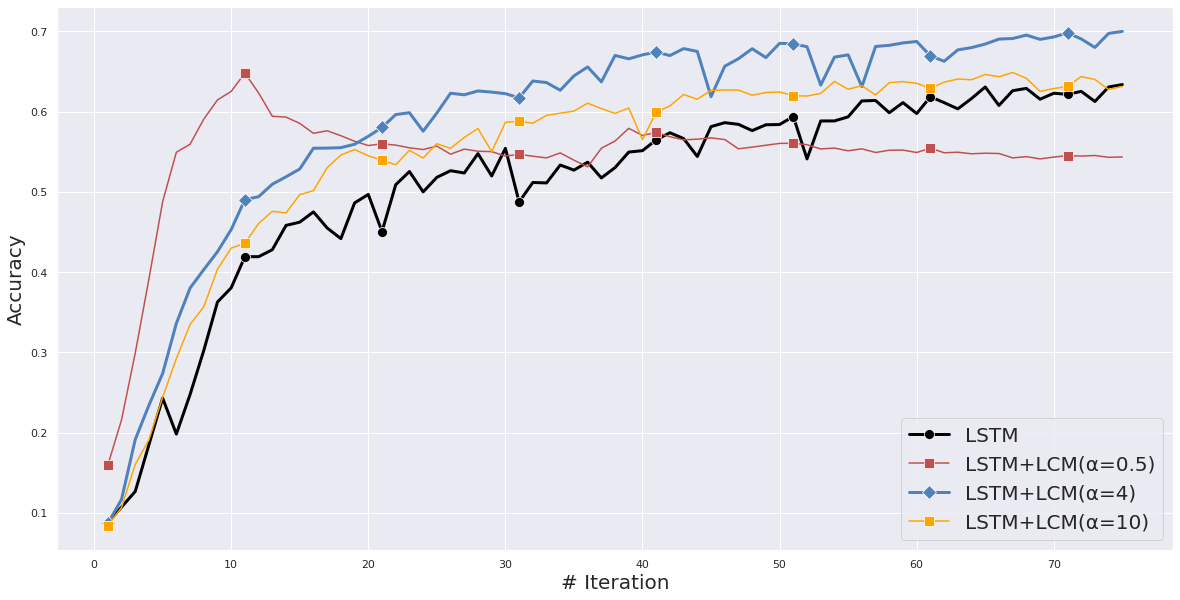}
         \caption{The effect of $\alpha$ in LCM}
         \label{AlphaEffect}
     \end{subfigure}
     \hfill
     \begin{subfigure}[t]{0.45\textwidth}
         \centering
         \includegraphics[width=\textwidth]{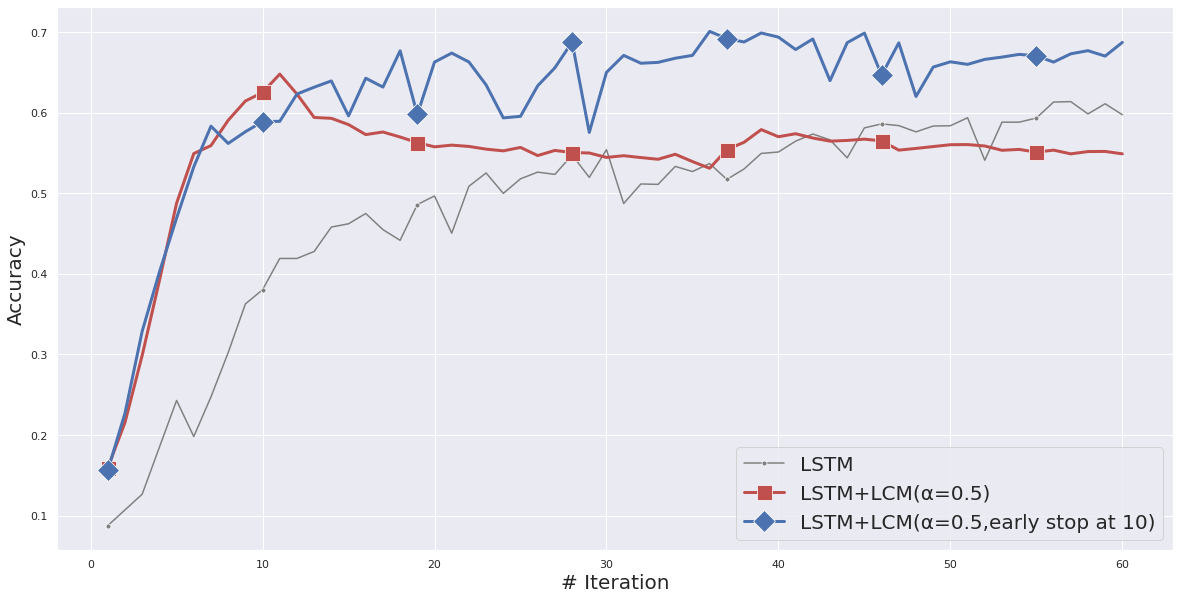}
         \caption{Early stop of LCM enhancement}
         \label{EarlyStop}
     \end{subfigure}
\caption{Hyper-parameter analysis of LCM-based models, including the effect of $\alpha$ and the early stop strategy for LCM enhancement.}
\label{LearningCurve}
\end{figure*}

\section{Experiments}
\subsection{Experiment Setup}

\subsubsection{Datasets}
To assess the effectiveness of our proposed method, we choose 5 benchmark datasets, including 3 English datasets and 2 Chinese datasets:

\noindent $\bullet$ The 20NG dataset\footnote{https://www.cs.umb.edu/\~ smimarog/textmining/datasets/} (bydata version) is an English news dataset that contains 18846 documents evenly categorized into 20 different categories.\\
$\bullet$  The AG’s News dataset\footnote{http://www.di.unipi.it/\~ gulli} is constructed by Xiang Zhang \cite{zhang2015character} which contains 127600 samples with 4 classes. We choose a subset of size 50000 samples in our experiments.\\
$\bullet$ The DBPedia dataset\footnote{http://dbpedia.org} is also created by Xiang Zhang \cite{zhang2015character}. It is an ontology classification dataset which has 630000 samples categorized into 14 classes. We randomly selected 50000 samples to form our experiment dataset.\\
$\bullet$ The FDCNews dataset\footnote{http://www.nlpir.org} is provided by Fudan University which contains 9833 Chinese news categorized into 20 different classes.\\
$\bullet$ The THUCNews dataset\footnote{http://thuctc.thunlp.org} is a Chinese news classification dataset collected by Tsinghua University. We constructed a subset from it which contains 39000 news evenly split into 13 news categories.

\subsubsection{Choice of Basic Predictors}
Our LCM is proposed as an enhancement for current main stream models, therefore, we only select some widely used model structures in text classification tasks.

\noindent $\bullet$ \textbf{LSTM}: We implement the LSTM model defined in  \cite{liu2016recurrent} which use the last hidden state as the text representation. We tried LSTM-rand which use random weights initialization for embedding layer and LSTM-pre that initialize with pre-trained word embeddings.\\
$\bullet$ \textbf{CNN}: We use the CNN structure in \cite{kim2014convolutional} and explore both CNN-rand and CNN-pre with and without using pre-trained word vectors respectively.\\
$\bullet$ \textbf{BERT}: Bidirectional Encoder Representations from Transformers \cite{devlin2018bert}. For faster training, we apply BERT-tiny \cite{bert-tiny} for English datasets. and ALBERT \cite{albert} for Chinese datasets.

\subsubsection{Settings}
For LSTM we set embedding size and hidden size as 64. For CNN, we use 3 filters with size 3, 10 and 25 and the number of filters for each convolution block is 100. For both LSTM and CNN models, the embedding size is 64 if no pre-trained word embedding are used. Otherwise, the embedding size is 250 for Chinese tasks and 100 for English tasks. The Chinese word embedding is pre-trained in around 1GB Chinese Wikipedia corpus using skip-gram \cite{mikolov2013distributed} algorithm. The English word embedding we choose is the GloVe embedding \cite{pennington2014glove}. In BERT models, we obtain text representations from the BERT model and then use a dense layer with 64 units to decrease the dimension of the text representation to 64. The LCM component in our case is implemented using an embedding lookup layer followed by a dense neural net, where the embedding size and the hidden size are kept the same as the basic predictor. The $\alpha$ decides the importance of LCM on the basic predictor. In our main experiments we just set $\alpha=4$ as a moderate choice. But by carefully tuned, the perfomance can get better. We train our model's parameters with the Adam Optimizer \cite{kingma2014adam} with an initial learning rate of 0.001 and batch size of 128. The model is implemented using Keras and is trained on GPU GeForce GTX 1070 Ti.

\begin{table*}[t]
\small
\centering
\begin{tabular}{l | l | l | l | l} 
 \toprule
& \textbf{8NG-H} & \textbf{8NG-E} & \textbf{4NG-H} & \textbf{4NG-E}\\
 \midrule
& 'comp.graphics', & 'comp.windows.x', & 'rec.autos', &        'comp.windows.x', \\
& 'comp.os.ms-windows.misc', & 'rec.sport.baseball', & 'rec.motorcycles', & 'rec.sport.baseball', \\
& 'comp.sys.ibm.pc.hardware', & 'alt.atheism',  & 'rec.sport.baseball', & 'alt.atheism', \\
\textbf{Labels} &    'comp.windows.x', & 'sci.med', & 'rec.sport.hockey' & 'sci.med' \\
& 'sci.crypt', & 'talk.politics.guns', &  &  \\
&  'sci.electronics', &  'misc.forsale', &  &  \\
&  'sci.med', & 'soc.religion.christian', &  &  \\
&  'sci.space' & 'talk.politics.misc' &  &  \\
\bottomrule
\end{tabular}
\caption{The labels of each 20NG subsets.}
\label{20NG_Subsets}
\end{table*}

\begin{table*}[t]
\small
\centering
\begin{tabular}{l | c c c c} 
 \toprule
 \textbf{Models} & \textbf{8NG-H} & \textbf{8NG-E} & \textbf{4NG-H} & \textbf{4NG-E} \\
 \midrule
 Basic Predictor & $0.7751$ & $0.8291 $ & $0.9048 $ & $\mathbf{0.9277}$\\
 Basic Predictor\ +\ LCM & $\mathbf{0.7973}$ & $\mathbf{0.8362}$ & $\mathbf{0.9147}$ & $0.9254 $\\ 
 \midrule
 \textit{Improvement} & $2.22\%$ & $0.72\%$ & $0.98\%$ & $-0.24\%$ \\
  \bottomrule
\end{tabular}
\caption{Test accuracy on some subsets of 20NG.}
\label{20NG_Subsets_Results}
\end{table*}

\subsection{Experimental Results}
Most of the datasets have already been split into train and test set. However the different split can directly affect the final performance of the model. Therefore, in our experiments, we combine the separated train and test set to one dataset and randomly split them to different train and test set 10 times by splitting ratio of 7:3. Then we assess all models 10 times on each the these different train test splitting. By doing this, we can better evaluate whether LCM can enhance the basic text classification predictors. 

\begin{table*}[t]
\small
\centering
\begin{tabular}{l | c  c c c} 
 \toprule
 \textbf{Models} & \textbf{Original 20NG} & \textbf{6\% Noise} & \textbf{12\% Noise} & \textbf{30\% Noise} \\
 \midrule
 LSTM & 0.6822 & 0.5946 & 0.5747 & 0.4681\\ 
 LSTM\ +\ LS & 0.7015 & 0.6155  & 0.6010 &  0.4994\\ 
 LSTM\ +\ LCM & \textbf{0.7242} & \textbf{0.6572} & \textbf{0.6385} & \textbf{0.5178}\\ 
 \midrule
 BERT & 0.8853 &  0.8695 &  0.8546 & 0.7916 \\ 
 BERT\ +\ LS & 0.8855 & 0.8742 &  0.8535 & 0.7932\\ 
 BERT\ +\ LCM & \textbf{0.8896} &  \textbf{0.8789} &  \textbf{0.8581} &  \textbf{0.7980} \\ 
 \bottomrule
 
\end{tabular}
\caption{Experiments on noisy datasets and comparison with label smoothing (LS) method, where the percentage means the proportion of samples randomly being mislabeled.}
\label{DataNoise}
\end{table*}

\subsubsection{Test Performance}
Table \ref{Results} presents the test performance and the improvement of LCM-based models compared with their corresponding basic predictors grouped by the structure. We can see from the results that LCM-based classification models outperform their baselines in all datasets when using LSTM-rand, CNN-rand and BERT structures. The LCM-based CNN-pre model is lightly worse in FDCNews and 20NG datasets. And in the LCM-based LSTM-pre model is not significantly different with the baseline in AG's News. The overall results in 5 datasets using 3 widely used deep learning structures illustrate that LCM has the ability to enhance the performance of text classification models. We can also see that LCM-base models usually have lower standard deviation, which reflects their robustness on the datasets splitting.

The biggest improvement was achieved by LCM on baseline LSTM-rand on the 20NG dataset, with 4.20\% increase in test accuracy. The performance gain on CNN-rand on the same dataset is also quite obvious with 1.04\% of improvement. The natural confusion of the labels in 20NG dataset can shine light on the reason why LCM performs quite well. Although there are 20 categories in 20NG dataset, these categories can naturally be divided into several groups. Therefore, it is natural that the labels in the same group are more difficult for model to distinguish. We further visualize the learned label representations of the 20 labels in 20NG dataset and is shown in Figure \ref{LabelEmbeddingVis}. The label  representations are extracted from the embedding layer of LCM. Figure \ref{LabelSimilarityMatrix} illustrate the cosine similarity matrix of the label representations, where the elements off the diagonal reflects how one label is similar to another label. Then we use t-SNE \cite{tsne} to visualize the high-dimensional representations on a 2D map, which is shown in Figure \ref{LabelTSNE}. We can find that the labels that are easily confused, especially those in the same group, tend to have similar representations. Since the label representations are all randomly initialized at the beginning, we can see that the LCM has the ability to learn very meaningful representations of the labels, which reflect the confusion between labels.

The reason why LCM-based classification models can usually get better test performance can be concluded into several aspects: (1) The LCM part learns the simulated label distribution SLD during training which captures the complex relation among labels by considering the semantic similarity between input document and labels. This is superior to simply using one-hot vector to represent the true label. (2) It is common that there exists some mislabeled data, especially for datasets with a large number of categories or very similar labels. In this scenario, training with one-hot label representation tend to be influenced by these mislabeled data more severely. However, by using SLD, the value on the index of the wrong label will be crippled and allocated to those similar labels. Therefore the misleading of the wrong label is relatively trivial. (3) Apart from the mislabeled data, when the given labels share some similarity (for example, ``computer" and ``electronics" are similar topics in semantics and share many key words in content), it is natural and reasonable to label a text sample with a label distribution that tells the various aspects of information. However, current classification learning paradigm ignore the difference between the samples and this important information is lost.

\subsubsection{The Effect of $\mathbf{\alpha}$ and Early Stop of LCM}
The $\alpha$ is a controlling hyper-parameter to decide the importance of LCM on the original model. The larger $\alpha$ will give the original one-hot label more weight when generating the SLD, thus reduce the influence of LCM. To ensure that the largest value in SLD keeps the same position with the original one-hot label, the $\alpha$ should be at least $0.5$. Figure \ref{AlphaEffect} shows the accuracy curve of different $\alpha$ on 20NG dataset using LSTM as basic structure. We can see from the graph, in the early stage, LCM-based models learn much faster than baseline model, and smaller $\alpha$ can lead to faster convergence. However, when $\alpha$ is too small, for example $\alpha=0.5$ in this case, the model will easily over-fit. In this situation, we can apply the early stop strategy on the LCM, that is, close the effect of the LCM influence at a certain number of iterations. From the leaning curve of $\alpha=0.5$ we find that the accuracy begin to decrease at about 10 epochs, then we can choose to stop LCM at about 10 iterations and continue train the model using the original one-hot label vector. The result shown in Figure \ref{EarlyStop} reveals the effectiveness of early stop of LCM enhancement that the model can prevent the over-fitting and continues to learn more to behave better. The choice of $\alpha$ and early stop strategy should be based on the nature of the specific dataset and task. According to the experience of our experiments, a smaller $\alpha$ tend to behave better in datasets with similar labels or labeling errors.



\subsection{The Influence of the Dataset Confusion Degree}
It is not easy to directly tell the confusion degree for each of our benchmark datasets, due to the difference in language, style and number of classes. Note that the 20NG dataset has some inner groups, which means some labels are naturally similar to each other in the same group. We sampled four subset from 20NG: 8NG-H, 8NG-E, 4NG-H and 4NG-E. The datasets with the ``H" are sampled from label groups so the documents are harder to distinguish, while the ``E" tag means the samples are selected from different groups and are easy to classify. The detailed labels of each datasets are listed in Table \ref{20NG_Subsets}. We choose LSTM as the basic predictor and the results on these four datasets are shown in Table \ref{20NG_Subsets_Results}. 
We can observe from the results that LCM helps a lot for 8NG-H and 4NG-H datasets but less helpful for 4NG-E, and even slightly decrease the accuracy for 4NG-E. It is straight-forward that 8NG-H has a higher confusion degree than 8NG-E, and so is to 4NG-H compared to 4NG-E. This phenomenon proves that LCM is especially helpful for datasets with high confusion degree.

\subsection{Experiments on Noisy Datasets and Comparison with Label Smoothing}
The success of deep learning models heavily depends on large annotated data, noisy data with labeling errors will severely diminish the classification performance which usually leads to an overfitted model. We constructed some noisy datasets with different percentage of noise from 20NG dataset, since 20NG inherently has some label groups. To make the noisy datasets closer to reality, the mislabeling samples are all chosen from the same label group, such as ``comp", ``rec" and ``talk". Then we conduct experiments to verify the effect of label smoothing (LS) and LCM. We explore two deep learning models as basic predictor. The results are shown in Table \ref{DataNoise}. The smoothing hyper-parameter is set to 0.1 for LS, and $\alpha =4$ for LCM. From the results we can see that LCM outperforms LS to a large degree, both in original cleaner dataset and datasets with obvious label errors.

\subsection{The Application of LCM on Images}
In fact, LCP is a common and natural problem in all classification tasks, not limited to text classification. For example, in the famous MNIST handwritten digits classification task, we can find that number ``0" looks similar to number ``6", and ``5" is similar to ``8" in many cases. Therefore, simply representing the label with a one-hot vector will omit these similarity information among labels. Thus, the idea of LCM might also help with this problem. We choose MNIST dataset and the Fashion MNIST dataset \cite{xiao2017fashion} to evaluate the effectiveness of LCM on image classification tasks. 

\begin{table}[t]
\small
\centering
\begin{tabular}{l | c c} 
 \toprule
 \textbf{Models} & \textbf{MNIST} & \textbf{Fashion MNIST} \\
 \midrule
 Basic Predictor & 0.9822 & 0.8929 \\ 

 Basic Predictor\ +\ LS & 0.9834 & 0.9009 \\ 

 Basic Predictor\ +\ LCM & \textbf{0.9841} & \textbf{0.9028} \\ 
 \bottomrule
\end{tabular}
\caption{Test accuracy on image classification tasks. Here the basic predictor is a simple CNN model.}
\label{ImageResults}
\end{table}

Due to the limited time and resource, we only implement a simple CNN network as the basic predictor. The results are shown in Table \ref{ImageResults}, which indicates  that LCM can also be used in image classification tasks to improve performance. More basic predictor networks and datasets will be experimented in the future.

\section{Conclusion and Future Work}
In this work, we propose Label Confusion Model (LCM) as an enhancement component to current text classification models to improve their performance. LCM can capture the relations between instances and labels as well as the dependency among labels. Experiments on five benchmark datasets proved LCM's enhancement on several popular deep learning models such as LSTM, CNN and BERT.

Our future work include the following directions: (\textit{i}) Designing a better LCM structure for computer vision tasks and conducting more experiments on image classification. (\textit{ii}) Generalizing the LCM method to multi-label classification problems and label distribution prediction.

\bibliography{ref.bib}
\end{document}